\documentclass{article} % For LaTeX2e
\usepackage{iclr2020_conference,times}

\usepackage{amsmath,amsfonts,bm}

\def\eqref#1{equation~\ref{#1}}
\def\1{\bm{1}}

\DeclareMathAlphabet{\mathsfit}{\encodingdefault}{\sfdefault}{m}{sl}
\SetMathAlphabet{\mathsfit}{bold}{\encodingdefault}{\sfdefault}{bx}{n}

\usepackage{hyperref}
\usepackage{url}

\usepackage[final]{graphicx}
\usepackage{booktabs}
\title{On Optimal Transformer Depth for Low-Resource Language Translation}

\author{Elan van Biljon \\
InstaDeep Ltd \& Stellenbosch University \\
Cape Town, South Africa \\
\texttt{e.vanbiljon@instadeep.com} \\
\And
Arnu Pretorius \\
InstaDeep Ltd \\
Cape Town, South Africa \\
\texttt{a.pretorius@instadeep.com} \\
\And
Julia Kreutzer \\
Heidelberg University \\
Heidelberg, Germany \\
\texttt{kreutzer@cl.uni-heidelberg.de}
}

\iclrfinalcopy % Uncomment for camera-ready version, but NOT for submission.
\begin{document}

\maketitle

\section{Introduction}
Transformers \citep{vaswani2017attention} have shown great promise as an approach to Neural Machine Translation (NMT) for low-resource languages \citep{abbott2018towards,martinus2019focus}. 
However, at the same time, transformer models remain difficult to optimize and require careful tuning of hyper-parameters to be useful in this setting \citep{popel2018training, nguyen2019transformers}. 
Many NMT toolkits come with a set of default hyper-parameters, which researchers and practitioners often adopt for the sake of convenience and avoiding tuning. These configurations, however, have been optimized for large-scale machine translation data sets with several millions of parallel sentences for European languages like English and French.

In this work, we find that the current trend in the field to use very large models is detrimental for low-resource languages, since it makes training more difficult and hurts overall performance, confirming the observations by \citet{murray2019auto,Fan2019ReducingTD}.
% Julia: what do you mean by 'hurting overall performance'? These models work quite well if there's lots of data, and even in multilingual transfer to low-resources. But for small data, they are probably overparametrized
% Elan: what we mean is that the networks are so deep that information about the input data is lost before it reaches the output. So even the transformers that are performing really well on languages with more data might do better if they were less deep.
% Julia: got it! But the trend in industry seems to be going in the opposite direction. Larger data -> larger models (and in industry hyperparameter tuning is done a lot). See BERT, GPT-2, multilingual NMT by Google. I wonder what needs to be changed to make those smaller models more successful!
Specifically, we compare shallower networks to larger ones on three translation tasks, namely: translating from English to Setswana, Sepedi (Northern Sotho), and Afrikaans. We achieve a new state-of-the-art BLEU score \citep{papineni2002bleu} on some tasks (more than doubling the previous best score for Afrikaans) when using networks of appropriate depth. Furthermore, we provide a preliminary theoretical explanation for this effect on performance as a function of depth. Overall, our findings seem to advocate the use of shallow-to-moderately sized deep transformers for NMT for low-resource language  translation.

Our intuition concerning the relationship between performance and depth stems from prior work on signal propagation theory in noise-regularised neural networks \citep{schoenholz2016deep, pretorius2018critical}. 
Specifically, \cite{pretorius2018critical} showed that using Dropout \citep{srivastava2014dropout} limits the depth to which information can stably propagate through neural networks when using ReLU activations.
Since both dropout and ReLU have been core components of the transformer since its inception \citep{vaswani2017attention}, this loss of information is likely to be taking place and should be taken into account when selecting the number of transformer layers.
Although the architecture of a transformer is far more involved than those analysed by \citet{pretorius2018critical}, the fundamental building blocks remain the same. 
Thus, in this paper, we make use of the above theoretical insights as a guide to our analysis of depth's influence on performance in transformers.

We see our work as complementary to the Masakhane project (``Masakhane'' means ``We Build Together'' in isiZulu.)\footnote{\url{https://github.com/masakhane-io/}}
In this spirit, low-resource NMT systems are now being built by the community who needs them the most. 
However, many in the community still have very limited access to the type of computational resources required for building extremely large models promoted by industrial research. 
Therefore, by showing that transformer models perform well (and often best) at low-to-moderate depth, we hope to convince fellow researchers to devote less computational resources, as well as time, to exploring overly large models during the development of these systems. 

\section{Results}
We trained networks of three depths: shallow (2 transformer layers---1 encoder layer and 1 decoder layer), medium (6 transformer layers---3 encoder and 3 decoder), and deep (12 transformer layers---6 encoder and 6 decoder---as is used in \cite{vaswani2017attention}), each on English to (1) Setswana, (2) Sepedi, and (3)  Afrikaans translation tasks. The model configurations, weights, and code are all available online at \url{https://github.com/ElanVB/optimal_transformer_depth}.

For comparability to previous work, all models were trained on the Autshumato data set \citep{autshumato} as it was preprocessed by \citet{martinus2019focus} and hyper-parameter settings have been left as similar as possible to previous work.

Table~\ref{tab:dataset} presents a breakdown of the relevant languages in the Autshumato data set \citep{martinus2019focus}.
This is done to contrast the small size when compared to many of the corpora that are represented at the Conference on Machine Translation (WMT) \citep{ng2019facebook}.

\begin{table*}[h]
	\begin{center}
        \begin{tabular}{l|ccc|cc}
    	    \toprule
    	    Data source & \multicolumn{3}{c}{Autshumato} & \multicolumn{2}{c}{WMT} \\
    	    \toprule
    	    Language & Setswana & Sepedi & Afrikaans & German & Russian \\
    % 		\hline
    		\midrule
            Total sentences & 123 868 & 30 777 & 53 172 & 27 700 000 & 26 000 000 \\
    	    \bottomrule
    	\end{tabular}
	\end{center}
	\caption{\emph{A breakdown of the number of sentences per language in the Autshumato data set compared to languages with higher global representation.} The number of sentences that can be used for training is multiple orders of magnitude less than many data sets presented at the Conference on Machine Translation (WMT).}
	\label{tab:dataset}
\end{table*}

\begin{figure}[h]
  \centering
  \includegraphics[width=0.6\linewidth]{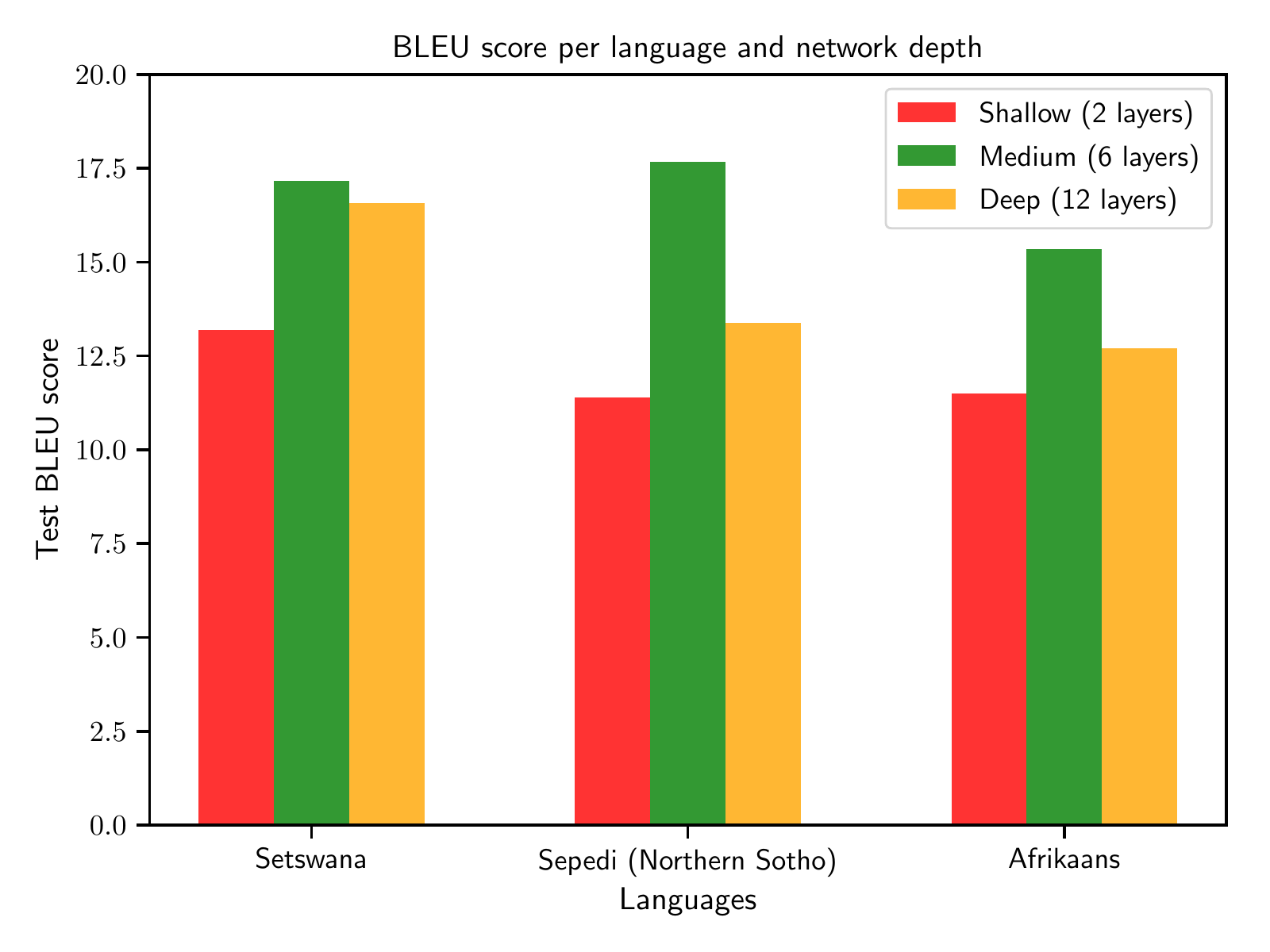}
  \caption{\emph{Networks of moderate depth perform better.} Test BLEU scores for networks of three depths: shallow (2 transformer layers), medium (6 transformer layers), and deep (12 transformer layers), each on English to (1) Setswana, (2) Sepedi, and (3) Afrikaans translation tasks.}
  \label{fig:depth_test}
\end{figure}

Figure~\ref{fig:depth_test} shows the quality of test set translations in terms of BLEU. We see that medium-depth models outperform deeper ones, thus we allowed the medium-depth networks to train for longer and report their performance (compared to previous work) in Table~\ref{tab:results}.
The medium-depth networks achieve higher scores for two of the three tasks (English to Setswana and English to Afrikaans translation) than the previous baselines as shown by the test BLEU scores in Table~\ref{tab:results}.
This is preliminary evidence showing that transformers consisting of 3 encoder and 3 decoder layers may outperform the canonical configuration \citep{vaswani2017attention} of 6 encoder and 6 decoder layers.

\begin{table*}[h]
	\begin{center}
        \begin{tabular}{l|cc}
    	    \toprule
    	    & Ours & \cite{martinus2019focus} \\
    		\toprule
            Setswana & \textbf{30.49} & 28.07 \\
            Sepedi (Northern Sotho) & 17.67 & \textbf{24.16} \\
            Afrikaans & \textbf{72.43} & 35.26 \\
    	    \bottomrule
    	\end{tabular}
	\end{center}
	\caption{\emph{Test BLEU scores for moderate depth networks compared to previous baselines.} Networks of moderate depth (6 layers in total, marked as ``Ours'') outperform previous baselines for Enlish to Setswana and Afrikaans translation.}
	\label{tab:results}
\end{table*}

Note that Table~\ref{tab:results} compares our results only to works that have been trained on the same data set and version thereof.
Whilst we are aware that better performing models exist for English to Setswana translation \citep{abbott2018towards,kato}, as best as we can tell those models are either trained on closed data sets or on a different version of the Autshumato data set.

Notably, our English to Afrikaans translation model more than doubles the previous BLEU baseline and seems to also outperform Statistical Machine Translation (SMT) \citep{daniel2014}. \footnote{However, the data set used by \citet{daniel2014} is not publicly available for a direct comparison.}

The BLEU metric with surface-based n-gram scoring might not be expressive enough for agglutinative languages like Sepedi and Setswana.
Therefore we also include example model outputs in Appendix~\ref{app:qual} for qualitative comparison.

Despite our networks of moderate depth outperforming previous Setswana and Afrikaans baselines, our Sepedi model performs significantly worse than the previous baseline.
It should be noted that we were unable to reproduce the baseline score obtained in \cite{martinus2019focus}.
Even so, our attempts to reproduce their result yielded models with very similar performance to our network of moderate depth with the two approaches usually being within approximately 0.5 test BLEU of each other.

\section{Discussion}
Our exploration into networks of moderate depth was largely due to preliminary signal propagation analyses we performed on simplified transformer layers.
However, we do not present these preliminary theoretical results here (left to be explored further in future work), but instead refer the reader to very recent and concurrent work done by \cite{bachlechner2020rezero} for a more complete motivation as well as a proposed solution.

Even though we do not achieve state-of-the-art results on all languages, we come very close, with approximately half the number of parameters and far less training time.
We believe there is still some room for improving the performance of our moderate-depth models by more carefully tuning their hyper-parameters.
However, we note that (1) finding stable learning rates can be very computationally expensive (and therefore be beyond what the community might currently be able to afford), and (2), in doing so, our work may become less comparable to those that have come before.

\bibliography{ms}

\begin{thebibliography}{16}
\providecommand{\natexlab}[1]{#1}
\providecommand{\url}[1]{\texttt{#1}}
\expandafter\ifx\csname urlstyle\endcsname\relax
  \providecommand{\doi}[1]{doi: #1}\else
  \providecommand{\doi}{doi: \begingroup \urlstyle{rm}\Url}\fi

\bibitem[Abbott \& Martinus(2018)Abbott and Martinus]{abbott2018towards}
Jade~Z Abbott and Laura Martinus.
\newblock Towards neural machine translation for african languages.
\newblock \emph{NeurIPS Workshop on Machine Learning for the Developing World},
  2018.

\bibitem[Bachlechner et~al.(2020)Bachlechner, Majumder, Mao, Cottrell, and
  McAuley]{bachlechner2020rezero}
Thomas Bachlechner, Bodhisattwa~Prasad Majumder, Huanru~Henry Mao, Garrison~W.
  Cottrell, and Julian McAuley.
\newblock Rezero is all you need: Fast convergence at large depth.
\newblock \emph{ArXiV}, abs/2003.04887, 2020.

\bibitem[Fan et~al.(2019)Fan, Grave, and Joulin]{Fan2019ReducingTD}
Angela Fan, Edouard Grave, and Armand Joulin.
\newblock Reducing transformer depth on demand with structured dropout.
\newblock \emph{ArXiv}, abs/1909.11556, 2019.

\bibitem[Groenewald \& du~Plooy(2010)Groenewald and du~Plooy]{autshumato}
J.~Hendrik Groenewald and Liza du~Plooy.
\newblock Processing parallel text corpora for three south african language
  pairs in the autshumato project.
\newblock In \emph{Proceedings of the Second Workshop on African Language
  Technology}, Valletta, Malta, 2010.

\bibitem[Martinus \& Abbott(2019)Martinus and Abbott]{martinus2019focus}
Laura Martinus and Jade~Z Abbott.
\newblock A focus on neural machine translation for african languages.
\newblock \emph{ArXiv}, abs/1906.05685, 2019.

\bibitem[Murray et~al.(2019)Murray, Kinnison, Nguyen, Scheirer, and
  Chiang]{murray2019auto}
Kenton Murray, Jeffery Kinnison, Toan~Q Nguyen, Walter Scheirer, and David
  Chiang.
\newblock Auto-sizing the transformer network: Improving speed, efficiency, and
  performance for low-resource machine translation.
\newblock In \emph{Proceedings of the 3rd Workshop on Neural Generation and
  Translation}, pp.\  231--240, 2019.

\bibitem[Ng et~al.(2019)Ng, Yee, Baevski, Ott, Auli, and
  Edunov]{ng2019facebook}
Nathan Ng, Kyra Yee, Alexei Baevski, Myle Ott, Michael Auli, and Sergey Edunov.
\newblock Facebook fair's wmt19 news translation task submission, 2019.

\bibitem[Nguyen \& Salazar(2019)Nguyen and Salazar]{nguyen2019transformers}
Toan~Q Nguyen and Julian Salazar.
\newblock Transformers without tears: Improving the normalization of
  self-attention.
\newblock \emph{International Workshop on Spoken Language Translation}, 2019.

\bibitem[Papineni et~al.(2002)Papineni, Roukos, Ward, and
  Zhu]{papineni2002bleu}
Kishore Papineni, Salim Roukos, Todd Ward, and Wei-Jing Zhu.
\newblock Bleu: a method for automatic evaluation of machine translation.
\newblock In \emph{Proceedings of the 40th Annual Meeting on Association for
  Computational Linguistics (ACL)}, Philadelphia, {PA, USA}, 2002.

\bibitem[Popel \& Bojar(2018)Popel and Bojar]{popel2018training}
Martin Popel and Ond{\v{r}}ej Bojar.
\newblock Training tips for the transformer model.
\newblock \emph{The Prague Bulletin of Mathematical Linguistics}, 110\penalty0
  (1):\penalty0 43--70, 2018.

\bibitem[Pretorius et~al.(2018)Pretorius, Biljon, Kroon, and
  Kamper]{pretorius2018critical}
Arnu Pretorius, Elan~Van Biljon, Steve Kroon, and Herman Kamper.
\newblock Critical initialisation for deep signal propagation in noisy
  rectifier neural networks, 2018.

\bibitem[Ronald \& Barnard(2007)Ronald and Barnard]{kato}
Kato Ronald and Etienne Barnard.
\newblock Statistical translation with scarce resources: a south african case
  study.
\newblock 98, 12 2007.

\bibitem[Schoenholz et~al.(2016)Schoenholz, Gilmer, Ganguli, and
  Sohl-Dickstein]{schoenholz2016deep}
Samuel~S Schoenholz, Justin Gilmer, Surya Ganguli, and Jascha Sohl-Dickstein.
\newblock Deep information propagation.
\newblock \emph{ArXiv}, abs/1611.01232, 2016.

\bibitem[Srivastava et~al.(2014)Srivastava, Hinton, Krizhevsky, Sutskever, and
  Salakhutdinov]{srivastava2014dropout}
Nitish Srivastava, Geoffrey Hinton, Alex Krizhevsky, Ilya Sutskever, and Ruslan
  Salakhutdinov.
\newblock Dropout: A simple way to prevent neural networks from overfitting.
\newblock \emph{Journal of Machine Learning Research}, 15\penalty0
  (1):\penalty0 1929--1958, 2014.

\bibitem[van Niekerk(2014)]{daniel2014}
Daniel~R. van Niekerk.
\newblock Exploring unsupervised word segmentation for machine translation in
  the south african context.
\newblock In \emph{Proceedings of Pattern Recognition Association of South
  Africa}, Cape Town, South Africa, 2014.

\bibitem[Vaswani et~al.(2017)Vaswani, Shazeer, Parmar, Uszkoreit, Jones, Gomez,
  Kaiser, and Polosukhin]{vaswani2017attention}
Ashish Vaswani, Noam Shazeer, Niki Parmar, Jakob Uszkoreit, Llion Jones,
  Aidan~N Gomez, {\L}ukasz Kaiser, and Illia Polosukhin.
\newblock Attention is all you need.
\newblock In \emph{Advances in Neural Information Processing Systems
  {(NeurIPS)}}, Long Beach, {CA, USA}, 2017.

\end{thebibliography}
\bibliographystyle{iclr2020_conference}

\newpage
\appendix
\section{Appendix: qualitative results}
\label{app:qual}

\begin{table*}[h]
	\begin{center}
        \begin{tabular}{p{1.5cm}|p{11cm}}
    	    \toprule
    	    Language & Text \\
    		\toprule
            English & Chapter 1 : Purpose of meetings \\
            Setswana & Kgaolo 1 : Maikaelelo a dikopano \\
            \toprule
            English & Mono-cropping is the agricultural practice of growing one single crop year after year . \\
            Setswana & Go jala korong ke tiragalo ya bolemirui ya go jwala dijwalwa tse di jwalwang ngwaga le ngwaga morago ga ngwaga . \\
            \toprule
            English & Some examples of these are round worms , lung worms , tape worm and liver fluke . \\
            Setswana & Dikao dingwe tsa dibokwana tse ke dibokwana , dibokwana tsa makgwafo , dibokwana le seedi sa leswe . \\
    	    \bottomrule
    	\end{tabular}
	\end{center}
	\caption{\emph{Example English to Setswana translations produced by the final model.}}
	\label{tab:tn}
\end{table*}

\begin{table*}[h]
	\begin{center}
        \begin{tabular}{p{1.5cm}|p{11cm}}
    	    \toprule
    	    Language & Text \\
    		\toprule
            English & These deaths set in motion escalating conflicts between residents and anyone seemed to be working with the state. \\
            Sepedi & Dipoloko tše di beilwego ka go dithulano tša go thibela magareng ga badudi le mang le mang di swanetše go šoma le mmušo. \\
            \toprule
            English & Past developments and activities externalised many real costs, displacing them onto environments and people as negative environmental and health and safety impacts. \\
            Sepedi & Ditlhabollo tša ast le ditiro tše di lego gona tša go ba le koketšo ye mentši, di di bea tikologo le batho ba tikologo le go hlokomologa maphelo le polokego. \\
            \toprule
            English & The aim of the operations is to address the contact crimes and housebreakings that escalated recently in the area. \\
            Sepedi & Maikemišetšo a tshepedišo a tshepedišo ke go rarolla bosenyi le go thuba ka dintlong tšeo di sa tšwago go di sa šongwa. \\
    	    \bottomrule
    	\end{tabular}
	\end{center}
	\caption{\emph{Example English to Sepedi (Northern Sotho) translations produced by the final model.}}
	\label{tab:nso}
\end{table*}

\begin{table*}[h]
	\begin{center}
        \begin{tabular}{p{1.5cm}|p{11cm}}
    	    \toprule
    	    Language & Text \\
    		\toprule
            English & if the number of persons nominated exceeds the number of vacancies to be filled , the president , after consulting the relevant profession , must appoint sufficient of the nominees to fill the vacancies , taking into account the need to ensure that those appointed represent the profession as a whole . \\
            Afrikaans & indien die getal persone wat benoem is meer is as die getal vakatures wat gevul moet word, moet die president die betrokke professie afsonderlik moet pleeg word, genoeg van die benoemdes aanstel om die vakatures te vul, met inagneming van die behoefte om te verseker dat diegene wat aangestel word die professie as geheel verteenwoordig. \\
            \toprule
            English & If CK documents were lodged separately, it will take three weeks to complete the process. \\
            Afrikaans & Indien die BK dokumente afgehandel is, sal dit drie weke duur om die proses te voltooi. \\
            \toprule
            English & STELLENBOSCH LOCAL MUNICIPALITY - SERVICES: TRAFFIC, ACCIDENTS \& ROAD SAFETY \\
            Afrikaans & STELLENBOSCH PLAASLIKE MUNISIPALITEIT - DIENSTE: VERKEER, ONGELUKKE EN PADVEILIGHEID \\
    	    \bottomrule
    	\end{tabular}
	\end{center}
	\caption{\emph{Example English to Afrikaans translations produced by the final model.}}
	\label{tab:afr}
\end{table*}

\end{document}